\documentclass[letterpaper]{article} 
\usepackage[draft]{aaai2026}  
\usepackage{times}  
\usepackage{helvet}  
\usepackage{courier}  
\usepackage[hyphens]{url}  
\usepackage{graphicx} 
\usepackage{natbib}  
\usepackage{caption} 
\usepackage{algorithm}
\usepackage{algorithmic}
\usepackage{graphicx}
\usepackage{amsmath}
\usepackage{amssymb}
\usepackage{booktabs}
\usepackage{subcaption}
\usepackage{multirow}
\usepackage{pifont}

\usepackage{newfloat}
\usepackage{listings}
\DeclareCaptionStyle{ruled}{labelfont=normalfont,labelsep=colon,strut=off} 
\floatstyle{ruled}
\newfloat{listing}{tb}{lst}{}
\floatname{listing}{Listing}
\title{PhysCorr: Dual-Reward DPO for Physics-Constrained Text-to-Video Generation with Automated Preference Selection}
\author{
    Peiyao Wang\textsuperscript{\rm 1},
    Weining Wang\textsuperscript{\rm 2},
    Qi Li\textsuperscript{\rm 2}
}
\affiliations{
    \textsuperscript{\rm 1}Beijing Institute of Technology\\
    \textsuperscript{\rm 2}Institute of Automation, Chinese Academy of Sciences\\
    wangpeiyao2026@ia.ac.cn, {weining.wang, qli}@nlpr.ia.ac.cn
}
\usepackage{bibentry}

\begin{document}
\nocopyright

\maketitle

\begin{abstract}
Recent advances in text-to-video generation have achieved impressive perceptual quality, yet generated content often violates fundamental principles of physical plausibility — manifesting as implausible object dynamics, incoherent interactions, and unrealistic motion patterns. Such failures hinder the deployment of video generation models in embodied AI, robotics, and simulation-intensive domains. To bridge this gap, we propose PhysCorr, a unified framework for modeling, evaluating, and optimizing physical consistency in video generation. Specifically, we introduce PhysicsRM, the first dual-dimensional reward model that quantifies both intra-object stability and inter-object interactions. On this foundation, we develop PhyDPO, a novel direct preference optimization pipeline that leverages contrastive feedback and physics-aware reweighting to guide generation toward physically coherent outputs. Our approach is model-agnostic and scalable, enabling seamless integration into a wide range of video diffusion and transformer-based backbones. Extensive experiments across multiple benchmarks demonstrate that PhysCorr achieves significant improvements in physical realism while preserving visual fidelity and semantic alignment. This work takes a critical step toward physically grounded and trustworthy video generation.
\end{abstract}


\section{Introduction}

Recent breakthroughs in text-to-video generation \cite{wan2025wan, chen2024videocrafter2, kong2024hunyuanvideo, yang2024cogvideox} have led to significant advances in producing high-quality, temporally coherent videos. These advancements are driven by large-scale diffusion-based architectures that model long-range dependencies and scale with vast datasets and computational resources. These models have set new benchmarks in visual fidelity, enabling the generation of dynamic video content that aligns with complex textual descriptions.
However, despite these impressive developments, a critical limitation persists: the failure to adhere to fundamental physical laws. Generated videos often exhibit physical inaccuracies, such as unrealistic object interactions, violations of fluid dynamics, and the distortion of 3D spatial relationships.
For example, in Figure 1(a), the waves generated by the model do not rebound significantly after splashing against the rocky cliffs, but instead continue to rise in a physically implausible manner.
In Figure 1(b), the knife fails to leave any marks on the meat after cutting it, which violates basic expectations of material interaction. 
These physical inconsistencies severely limit the utility of these models in domains that require strict adherence to physical realism, such as scientific visualization and robotics.


\begin{figure}[t!] 
    \centering 
    \begin{subfigure}{0.47\textwidth}
        \centering
        \includegraphics[width=\linewidth]{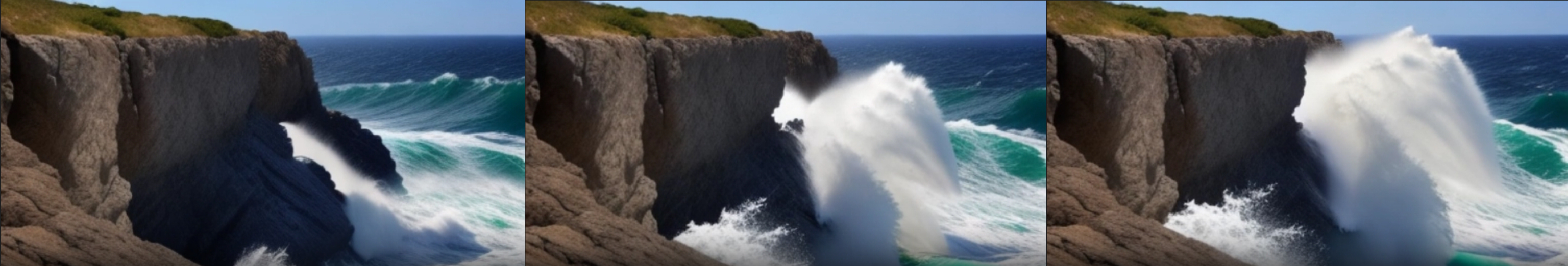}
        \caption{Video Generated by "Big waves splashing on rocky cliffs"}
        \label{fig:unreasonable-a}
    \end{subfigure}
    \hfill 
    \begin{subfigure}{0.47\textwidth}
        \centering
        \includegraphics[width=\linewidth]{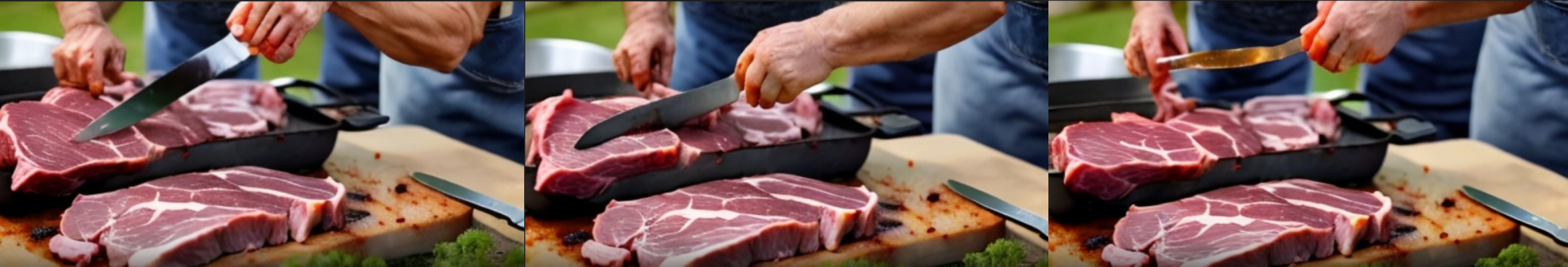}
        \caption{Video Generated by "Preparing meat for barbecue"}
        \label{fig:unreasonable-b}
    \end{subfigure}

    \caption{The videos generated by VideoCrafter2 using (a) "Big waves splashing on rocky cliffs" and (b) "Preparing meat for barbecue". }
    \label{fig:combined_events}
\end{figure}
Existing solutions \cite{prabhudesai2024video, yuan2024instructvideo} primarily focus on aligning generated videos with human preferences using techniques like Reinforcement Learning from Human Feedback (RLHF) and Direct Preference Optimization (DPO) \cite{liu2025videodpo}. 
These methods, while effective in enhancing visual quality and text-video coherence, fail to address the core issue of physical accuracy. 
These errors arise from three core limitations in existing reinforcement learning (RL) alignment frameworks. 
First, current reward models \cite{liu2025improving, wu2024boosting} rely heavily on large-scale VLMs, making training difficult and requiring a long time to evaluate video quality. Meanwhile, reward models prioritize frame-level aesthetics and text alignment, neglecting physical plausibility. 
Methods like VideoReward \cite{liu2025improving} and VIDEORM \cite{wu2024boosting} optimize for visual-textual correspondence but lack mechanisms to quantify physical plausibility, such as rigid body dynamics and fluid continuity. 
Second, human preference datasets focus on subjective quality dimensions, such as visual appeal and motion smoothness, while overlooking annotations for physical violations. This results in a misalignment between training objectives and real-world physical constraints. Third, standard alignment algorithms like DPO and RWR struggle to correct fundamental physical inaccuracies. Their KL-divergence constraints often preserve low-level artifacts, and inference-time guidance (e.g., Flow-NRG \cite{liu2025improving}) fails to resolve structural violations without dedicated physics-aware rewards.

To address these issues, we propose PhysCorr, a novel framework for physics-constrained text-to-video generation via structured preference learning. 
Our approach integrates two key innovations: \textit{Physics Reward Model (PhysicsRM)}: A lightweight evaluator combining subject-consistency (geometric stability, material integrity) and mechanical coherence (trajectory continuity, collision physics) to compute granular PhyScores. By distilling knowledge from a 7B-parameter VLM to a 0.5B model via task-specific fine-tuning, PhysicsRM achieves parameter efficiency while retaining robustness in physical reasoning. \textit{Physics-Specialized DPO (PhyDPO)}: An automated alignment paradigm where PhyScores guide win/lose video pairing, and a reweighted DPO loss prioritizes high-delta physics violations (e.g., momentum non-conservation). This approach adaptively optimizes physical plausibility while preserving visual quality.

Extensive evaluations demonstrate that PhysCorr achieves state-of-the-art physics alignment across multiple benchmarks. When augmented with leading models like Wan2.1 and VideoCrafter2, it significantly enhances physical realism without compromising visual fidelity, addressing the core limitations of current RL-driven methods.

Our main contributions are as follows:
\begin{itemize}
\item  We introduce PhysCorr, a novel framework for physics-constrained text-to-video generation, combining PhysicsRM and PhyDPO to improve physical plausibility and visual fidelity while preserving realism.
\item We propose PhysicsRM, the first parameter-efficient reward model that quantifies physical plausibility by evaluating both subject-consistency (e.g., geometric stability, material integrity) and mechanical coherence (e.g., trajectory continuity, collision physics). Moreover, we introduced Huber loss for the first time in the training of video reward models. 
\item We perform extensive evaluations, demonstrating that PhysCorr significantly improves physical plausibility and visual quality in leading text-to-video models, outperforming state-of-the-art methods.
\end{itemize}

\section{Related Works}

\paragraph{Text-to-video Diffusion Models}
Text-to-video (T2V) generation has emerged as a dynamic frontier in AI, driven by advances in diffusion models \cite{sohl2015deep, karras2017progressive, ho2020denoising, song2020score, song2020denoising, blattmann2023stable, ding2021cogview, ho2022imagen, ding2022cogview2, huang2023collaborative, mou2024t2i}, variational autoencoder-based compression techniques \cite{kingma2013auto, van2017neural, esser2021taming, yu2023magvit, podell2023sdxl} and transformer architectures \cite{dosovitskiy2020image, podell2023sdxl}. Early T2V models synthesized primarily short clips (2-3 seconds) using diffusion frameworks \cite{blattmann2023align, chai2023stablevideo, ge2023preserve, guo2023animatediff, khachatryan2023text2video, luo2023videofusion, polyak2024movie, wang2025lavie, zhang2024show, zhou2022magicvideo, hong2022cogvideo} such as VideoCrafter \cite{chen2023videocrafter1}, Modelscope \cite{wang2023modelscope}, and nondiffusion alternatives (e.g. EMU \cite{dai2023emu, wang2024emu3}), focusing on improving visual fidelity and temporal consistency. These models often relied on U-Net or Diffusion Transformer (DiT) architectures (e.g., Open-Sora \cite{zheng2024open}, CogVideoX \cite{yang2024cogvideox}, Wan \cite{wan2025wan}). However, challenges persisted due to data complexity, leading to outputs that frequently fell short of user expectations in quality and text alignment. To address these limitations, post-training enhancement methods—including parameter-efficient tuning \cite{he2023scalecrafter, guo2024make, li2024t2v, li2024t2v2}, data-centric optimization \cite{he2024venhancer}, and human preference alignment \cite{prabhudesai2024video, yuan2024instructvideo}—were developed. A paradigm shift occurred with foundational models like Sora \cite{Sora}, which scaled training to unprecedented levels, demonstrating robust spatiotemporal coherence and enabling longer, higher-quality video synthesis. \cite{agarwal2025cosmos, fan2025vchitect, peng2025open, si2025repvideo, kong2024hunyuanvideo} This progress redirected focus toward intrinsic faithfulness: adherence to physical laws \cite{agarwal2025cosmos, Sora}, commonsense reasoning, and complex compositional integrity to support applications like AI filmmaking and world simulation.

\paragraph{Physical Plausibility Evaluation}
PhyGenBench \cite{meng2024towards}, the first benchmark specifically designed to quantify the physical rationality of video generative models, evaluates a model’s understanding of physical laws through VLMs. The recently released VBench2 \cite{zheng2025vbench} benchmark not only quantifies physical rationality related attributes, but also includes four other dimensions lying in intrinsic faithfulness - Human Fidelity, Controllability, Creativity, and Commonsense. It introduces a dedicated Physics dimension comprising three granular sub-abilities: State Change - Mechanics, State Change - Thermotics, and State Change - Material.

\begin{figure*}[h!]
    \centering
    \begin{subfigure}{0.72\textwidth}
        \includegraphics[height=6.45cm]{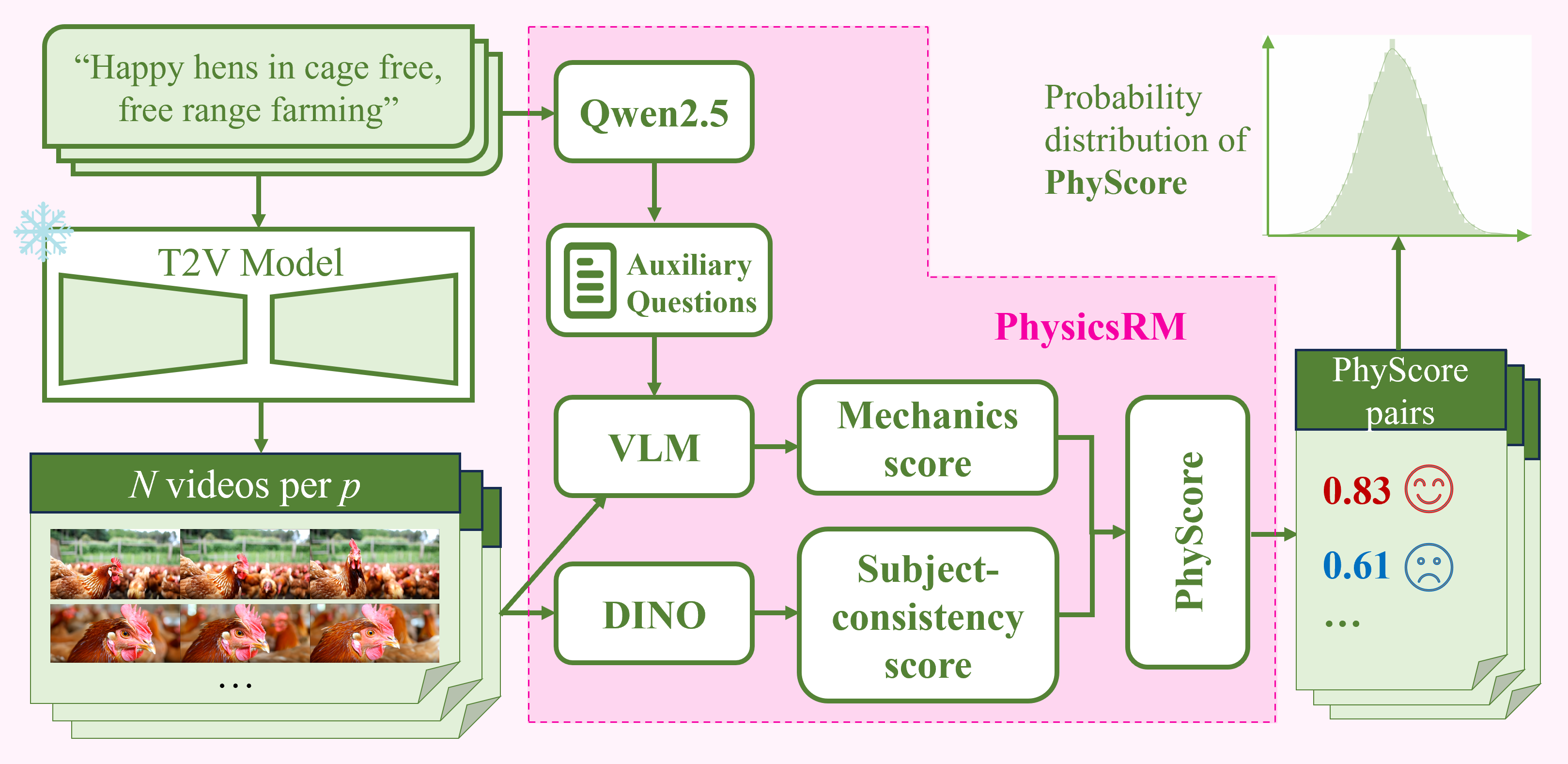} 
        \caption{}
        \label{fig:PhysicsRM}
    \end{subfigure}
    \hfill 
    \begin{subfigure}{0.26\textwidth}
        \includegraphics[height=6.45cm]{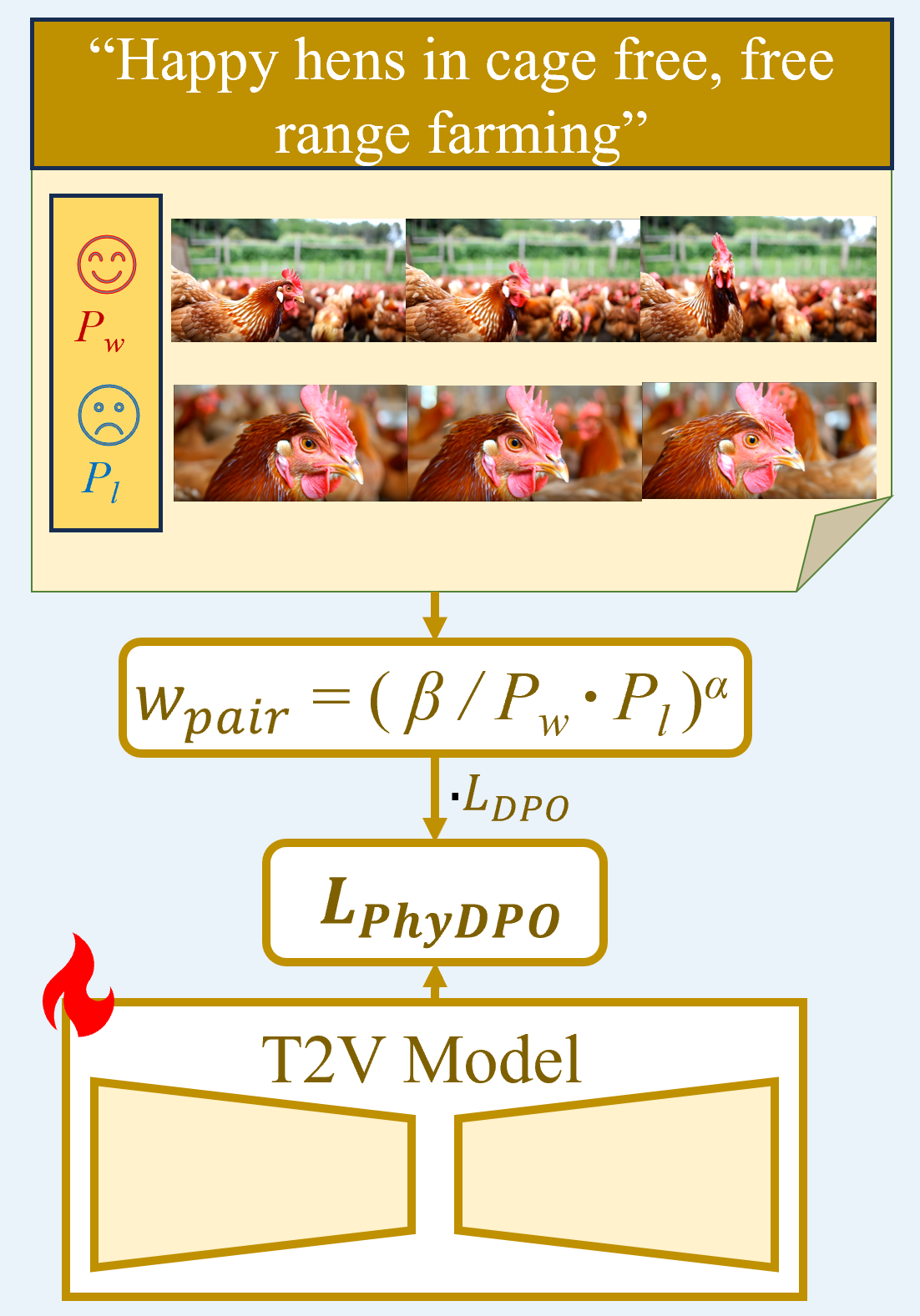} 
        \caption{}
        \label{fig:PhyDPO}
    \end{subfigure}
    
    \caption{\textbf{PhysCorr pipeline.} We propose (a) PhysicsRM integrating subject-consistency module and mechanics module to quantify physical plausibility (\textit{bottom}). For each prompt $p$, we generate $N$ videos using the target video diffusion model and compute their PhysicsRM-derived PhyScores. The highest-scoring video (physically plausible) and lowest-scoring video (physically implausible) form a preference pair for training. During (b) PhyDPO training, preference pairs are reweighted based on their PhyScore differance. Pairs with larger PhyScore difference (highlighting severe physical errors) receive higher weights, forcing the model to prioritize correcting egregious physical inaccuracies.}
    \label{fig:arch}
\end{figure*}

\paragraph{Preference Learning for Generative Models}
Reinforcement learning with human or AI feedback (RLHF/RLAIF) has become instrumental for aligning text-to-video (T2V) models with human preferences, primarily following two paradigms: leveraging VLMs as evaluators \cite{furuta2024improving, liu2025videodpo} or training specialized video reward models \cite{liu2025improving, wu2024boosting}. VLM-based methods assess the alignment between the generated videos and text descriptions, offering a comprehensive semantic analysis of the generated content. For reward-model-based approaches, they involve training a reward model (RM) on human-annotated winlose pairs to predict preference labels, followed by policy alignment via reinforcement learning. Although VLM-based methods offer flexibility in evaluation, trained RMs tend to provide more detailed video-specific understanding, though they require extensive data and training resources.

\section{Method}
In this section, we propose \textit{PhysCorr}, a comprehensive physics-aligned video generation pipeline through specialized DPO, as shown in Figure 2. Our pipeline introduces \textit{PhysicsRM}, which integrates a subject-consistency module and a mechanics module to evaluate the physical plausibility of generated videos. For each prompt, the model generates multiple videos and computes their \textit{PhyScores} based on physical plausibility. The video with the highest score is considered physically plausible, while the one with the lowest score is physically implausible, forming a preference pair for training. During \textit{PhyDPO} training, these preference pairs are weighted according to the \textit{PhyScore}. Pairs with a larger score difference are given higher weight, which helps the model focus on correcting severe physical inaccuracies.

\subsection{Lightweight Physics Reward Model}
This subsection introduces Lightweight Physics Reward Model (PhysicsRM), a novel reward model that quantifies physical plausibility by integrating geometric stability (subject-consistency) and mechanical verification. The subject-consistency module evaluates temporal feature coherence via DINOv2 embeddings, while the mechanics module employs a distilled LLaVA-Video-Qwen2 model for hierarchical physical reasoning. PhysicsRM is trained on human-annotated datasets using Huber loss, dynamically balancing both components via a trainable weight parameter $\lambda$ to optimize robustness against annotation noise and physical complexity.

The final PhyScore is computed as a weighted sum of these two components, where the weights are determined by a trainable parameter: 

\begin{equation}
s_{\text{phy}} = \sigma(\lambda) \cdot s_{\text{subj}} + (1 - \sigma(\lambda)) \cdot s_{\text{mech}}
\end{equation}
where $\lambda \in \mathbb{R}$ is a trainable parameter initialized at 0 (equivalent to $\sigma(0) = 0.5$), and $\sigma(\cdot)$ is the sigmoid function ensuring weight normalization.

\paragraph{Subject-Consistency Module}
This module measures the temporal stability of 3D geometric features across frames in the video. By extracting features using DINOv2 and computing cosine similarity between consecutive frames, it captures how consistent the video is with respect to its subject. The score is then normalized to improve the model’s robustness against varying video characteristics.

Given a video $\mathcal{V} = \{I_1, I_2, \dots, I_F\}$ with $F$ frames: firstly, extract DINOv2 features: $\phi_t = \text{DINO}(I_t) \in \mathbb{R}^d$; secondly, compute temporal consistency metric:
    \begin{equation}
    s_{\text{subj}} = \frac{1}{F-1} \sum_{t=1}^{F-1} \cos(\phi_t, \phi_{t+1})
    \end{equation}
    where $\cos(\cdot,\cdot)$ denotes cosine similarity; thidly, normalize score: $s_{\text{subj}} \leftarrow \frac{s_{\text{subj}} - \mu_{\text{subj}}}{\sigma_{\text{subj}}}$ using dataset statistics.

\paragraph{Mechanics Verification Module}
The mechanics verification module evaluates the mechanical plausibility of the video through a hierarchical two-stage process. It generates auxiliary questions related to the physics of the video and checks the answers using a lightweight distillation of the LLaVA-Video-Qwen2 model. The process ensures that videos are evaluated for mechanical correctness while maintaining computational efficiency through model distillation.

The two-stage hierarchical evaluation:
\begin{algorithm}
\caption{Mechanics Scoring Pipeline}
\label{alg:mechanics_scoring}
\begin{algorithmic}[1]
\REQUIRE Video $\mathcal{V}$, text prompt $p$
\STATE $q_1, q_2 \leftarrow \text{Qwen2.5}(p)$ \COMMENT{Generate auxiliary physics questions}
\STATE $\mathcal{A}_1 \leftarrow \text{LLaVA-Video-Qwen2-Distill}(\mathcal{V}, q_1)$ \COMMENT{Lightweight distilled model}
\IF{$\mathcal{A}_1$ incorrect}
    \STATE $s_{\text{mech}} \leftarrow 0$ \COMMENT{First-level failure}
\ELSE
    \STATE $\mathcal{A}_2 \leftarrow \text{LLaVA-Video-Qwen2-Distill}(\mathcal{V}, q_2)$
    \STATE $s_{\text{mech}} \leftarrow \begin{cases} 
        1 & \text{if } \mathcal{A}_2 \text{ correct} \\
        0.5 & \text{otherwise}
    \end{cases}$
\ENDIF
\RETURN $s_{\text{mech}}$
\end{algorithmic}
\end{algorithm}

Model Architecture: The \textit{LLaVA-Video-Qwen2-Distill} model is a lightweight video understanding system developed for efficient physical analysis. To minimize computational requirements while maintaining accuracy, we apply \textbf{model distillation} to the base \textit{LLaVA-Video-Qwen2-7B} architecture using the first two stages of the \textit{LLaVA-MoD} framework for efficient knowledge transfer. This distillation process transfers knowledge from the larger 7B parameter model to a compact variant with only 0.5B parameters, achieving 98\% of the original accuracy while reducing inference latency by 8.9$\times$.

The physics-oriented question generator employs a controlled text generation strategy. For a given prompt $p$, it produces a question pair $(q_1, q_2)$ through constrained decoding:

\begin{equation}
(q_1, q_2) \sim P_{\text{gen}}(q|p; \mathcal{C})
\label{eq:question_gen}
\end{equation}

where $P_{\text{gen}}$ is Qwen2.5's conditional language model, and $\mathcal{C}$ denotes the generation constraints:
\begin{align*}
\mathcal{C} = \{ & \texttt{difficulty}(q_1) < \texttt{difficulty}(q_2), \\
& \texttt{domain}(q_i) = \texttt{mechanics}, \\
& \texttt{relevance}(q_i, p) > \tau \}
\end{align*}

\paragraph{Training Prompts Collection}
We constructed our training prompts set based on the \textit{T2VQA} dataset. Our training set $\mathcal{D}_{\text{train}}$ is constructed as:

\begin{equation}
\mathcal{D}_{\text{train}} = \underbrace{\mathcal{D}_{\text{physics}}}_{\substack{\text{50 prompts} \\ \text{(difficult content)}}} \cup \underbrace{\mathcal{D}_{\text{non-physics}}}_{\substack{\text{250 prompts} \\ \text{(neutral content)}}}
\end{equation}

where: $\mathcal{D}_{\text{physics}}$ contains 50 manually selected prompts that are difficult to generate in terms of physical rationality, $\mathcal{D}_{\text{non-physics}}$ contains 250 randomly selected neutral prompts (1:5 ratio). 

\paragraph{Model Training}
PhysicsRM is trained via supervised learning with human-annotated physical plausibility scores. 
The training set contains every prompt in $\mathcal{D}_{\text{train}}$, and each prompt has 10 videos with annotated human $s_{\text{phy}} \in [0,1]$.

To mitigate annotation noise, the Huber loss is employed in model training, combining quadratic behavior for small errors and linear growth for large errors, which enabling precise learning and reducing outlier sensitivity: 

\begin{equation}
\mathcal{L}_{\text{PhysicsRM}} = \frac{1}{|\mathcal{D}|} \sum_{(p,\mathcal{V}) \in \mathcal{D}} \ell_{\delta}(s_{\text{phy}}^{\text{pred}} - s_{\text{phy}}^{\text{gt}})
\end{equation}
where $\ell_{\delta}(z) = \begin{cases} 
\frac{1}{2}z^2 & |z| \leq \delta \\
\delta(|z| - \frac{1}{2}\delta) & \text{otherwise}
\end{cases}$ with $\delta=0.2$.

\subsection{Win/Lose Video Selection}

This section outlines the process for selecting ``win" and ``lose" videos based on the PhyScores predicted by PhysicsRM. It involves curating physically challenging prompts, generating videos for each prompt, and assigning scores to each video to determine the best and worst-performing videos. The final dataset consists of pairs of videos labeled as ``win" or ``lose" based on their respective scores.

\textbf{Prompt Curation.} Firstly, select physically challenging prompts: 36 manually selected from \textit{Vidpro-10k} based on physical complexity criteria; secondly, get random prompts: 72 randomly sampled from the same dataset; thirdly, final composition: $1:2$ ratio of challenging vs. random prompts ($N_{\text{total}} = 108$).
    
\textbf{Video Generation.} 
For each prompt $p_i$, generate 4 videos using a pre-trained text-to-video model:
\[
\mathcal{V}_{i} = \{v_{i,j}\}_{j=1}^{8} \sim \text{Pre-trained T2V}(p_i)
\]
    
\textbf{Preference Pair Assignment.} 
PhysicsRM assigns scores $s_{i,j}$ to each video. For each prompt set:
\begin{equation}
\begin{aligned}
v^{\text{win}}_i= \underset{j}{\arg\max}  s_{i,j}, v^{\text{lose}}_i= \underset{j}{\arg\min}  s&_{i,j}, \\
\mathcal{D}_{\text{pref}}= \left\{ \left(p_i, v^{\text{win}}_i, v^{\text{lose}}_i \right) \right\}_{i=1}^{108}
\end{aligned}
\end{equation}

\subsection{Physics-Specialized DPO}

Traditional DPO training treats all preference pairs equally, regardless of the magnitude of quality differences between samples. However, we observe that physically indiscernible pairs (i.e., where the PhyScore difference $\Delta s = s^+ - s^-$ is minimal) provide weak learning signals during optimization. To address this, we propose the Physics-Specialized DPO (PhyDPO) that prioritizes high-impact pairs based on two physical principles. Firstly, pairs with larger PhyScore gaps reflect clearer physical superiority. Secondly, samples with extreme scores (very high/low) occur infrequently but are critical for constraint satisfaction.

The PhyScore-driven re-weighting pipeline operates in three stages as follows.

\textbf{Probability Density Modeling.} For $K \times N$ generated simulations (across $K$ prompts), we compute the empirical probability density $\hat{p}(s)$ of PhyScore $s$ using histogram binning with width $\delta = 0.01$:
\begin{equation}
    \hat{p}(s) = \frac{1}{K N \delta} \sum_{i=1}^{K N} \mathbf{1}\left[s - \frac{\delta}{2} \leq s_i < s + \frac{\delta}{2}\right]
\end{equation}

\begin{table*}[h!]
\centering
\label{tab:main_results}
\renewcommand{\arraystretch}{1.25}
\resizebox{1\textwidth}{!}{%
\begin{tabular}{l|c|cccccccc} 
\hline
Model & Total & 
\multicolumn{1}{c}{Subject} & 
\multicolumn{1}{c}{Background} & 
\multicolumn{1}{c}{Temporal} & 
\multicolumn{1}{c}{Motion} & 
\multicolumn{1}{c}{Dynamic} & 
\multicolumn{1}{c}{Aesthetic} & 
\multicolumn{1}{c}{Imaging} & 
\multicolumn{1}{c}{Object} \\
& & Consistency & Consistency & Flickering & Smoothness & Degree & Quality & Quality & Class \\ 
\hline
VideoCrafter2 & 82.08 & 96.85 & 98.22 & 98.41 & 97.73 & \textbf{42.50} & 63.13 & 67.22 & 92.55 \\
\textbf{PhysCorr (Ours)} & \textbf{82.31 (+0.28\%)} & \textbf{97.14 (+0.30\%)} & \textbf{98.60 (+0.39\%)} & \textbf{98.44 (+0.03\%)} & \textbf{97.92 (+0.19\%)} & 41.67 (-0.20\%) & \textbf{63.66 (+0.84\%)} & \textbf{67.84 (+0.92\%)} & \textbf{93.24 (+0.75\%)} \\
\hline
Wan2.1 & 88.36 & 95.92 & 97.39 & \textbf{99.53} & 96.92 & 94.35 & 61.53 & 67.28 & 94.24 \\
\textbf{PhysCorr (Ours)} & \textbf{88.63 (+0.31\%)} & \textbf{96.79 (+0.91\%)} & \textbf{97.45 (+0.06\%)} & 99.36 (-0.17\%) & \textbf{97.24 (+0.33\%)} & \textbf{94.80 (+0.48\%)} & \textbf{61.98 (+0.73\%)} & \textbf{67.32 (+0.06\%)} & \textbf{94.92 (+0.72\%)} \\
\hline
\end{tabular}%
}

\resizebox{1\textwidth}{!}{%
\begin{tabular}{l|c|cccccccc} 
\hline
Model & Total & 
\multicolumn{1}{c}{Multiple} & 
\multicolumn{1}{c}{Human} & 
\multicolumn{1}{c}{Color} & 
\multicolumn{1}{c}{Spatial} & 
\multicolumn{1}{c}{Scene} & 
\multicolumn{1}{c}{Appearance} & 
\multicolumn{1}{c}{Temporal} & 
\multicolumn{1}{c}{Overall} \\
& & Objects & Action &  & Relationship &  & Style & Style & Consistency \\ 
\hline
VideoCrafter2 & \textbf{49.87} & \textbf{40.66} & \textbf{95.00} & 92.92 & \textbf{35.86} & 55.29 & 25.13 & 25.84 & 28.23 \\
\textbf{PhysCorr (Ours)} & 49.75 (-0.24\%) & 40.24 (-1.03\%) & \textbf{95.00} & \textbf{92.98 (+0.06\%)} & 34.57 (-3.60\%) & \textbf{55.67 (+0.69\%)} & \textbf{25.15 (+0.08\%)} & \textbf{26.07 (+0.89\%)} & \textbf{28.34 (+0.39\%)} \\
\hline
Wan2.1 & 59.64 & 81.44 & \textbf{98.80} & 87.79 & \textbf{81.08} & 53.67 & \textbf{21.13} & 25.69 & 27.44 \\
\textbf{PhysCorr (Ours)} & \textbf{59.75 (+0.18\%)} & \textbf{82.19 (+0.92\%)} & 98.49 (-0.31\%) & \textbf{87.85 (+0.07\%)} & 80.36 (-0.89\%) & \textbf{53.84 (+0.32\%)} & 21.10 (-0.14\%) & \textbf{26.32 (+2.45\%)} & \textbf{27.82 (+1.38\%)} \\
\hline
\end{tabular}%
}
\caption{Comprehensive VBench evaluation of PhysCorr alignment. PhysCorr improves 25/32 dimensions across both models, with most significant gains in temoral style  and overall consistency.}
\end{table*}
    
\textbf{Weight Assignment.} For each preference pair $(v_i^+, v_j^-)$ with scores $(s_i^+, s_j^-)$, we define its \textit{joint sampling probability} as:
\begin{equation}
    \mathcal{P}_{ij} = \hat{p}(s_i^+) \cdot \hat{p}(s_j^-)
\end{equation}
The re-weighting coefficient is then computed as:
\begin{equation}
    w_{ij} = \left( \frac{\beta}{\mathcal{P}_{ij}} \right)^\alpha = \left( \frac{\beta}{\hat{p}(s_i^+)\hat{p}(s_j^-)} \right)^\alpha
\end{equation}
where $\beta = \max_s \hat{p}(s)$ normalizes the scale, and $\alpha > 0$ controls prioritization intensity.
    
\textbf{Physics-Informed Loss Integration.} The final training objective combines DPO loss with our re-weighting:
\begin{equation}
    \mathcal{L}_{\text{PhyDPO}} = \frac{1}{|\mathcal{D}|} \sum_{(p, v_i^+, v_j^-) \in \mathcal{D}} w_{ij} \cdot \mathcal{L}_{\text{DPO}}(p, v_i^+, v_j^-)
    \label{eq:phydpo_loss}
\end{equation}
where $\mathcal{L}_{\text{DPO}}$ is the standard DPO loss and $\gamma$ is the temperature hyperparameter.

\section{Experiments}
\subsection{Experiment Setup}
\textbf{Baselines.} We compare PhysCorr against two categories of baselines: (1) State-of-the-art open-source video generative models: VideoCrafter2, an innovative U-Net architecture video generative model that efficiently creates high-quality videos using low-quality video and high-quality image data by decoupling visual and motion information, and Wan2.1, a cutting-edge Diffusion Transformers (DiT) architecture video foundation model that generates high-quality videos from text and images, supporting multiple generation tasks; (2) Ablated versions of our framework: PhysicsRM$_{\text{single}}$ (using only subject-consistency) and PhyDPO$_{\text{unweighted}}$ (without reweighting).

\noindent\textbf{Benchmarks.} We use two benchmark suites for evaluation: VBench, a widely recognized benchmark to evaluate the quality and semantic consistency of video generation across 16 hierarchical dimensions, providing fine-grained assessment. VBench2, introducing the first video generation benchmarking suite with physics compliance metrics, evaluating intrinsic faithfulness beyond visual quality.

\noindent\textbf{Implementation Details.} We augment two base models: VideoCrafter2 and Wan2.1-14B. For each prompt, we generate $N=4$ videos. PhysicsRM uses LLaVA-Video-Qwen2-Distill (0.5B params) with $\lambda$ initialized at 0.5. Training uses AdamW optimizer ($lr=5e-6$), batch size 4, for 2K steps on 4$\times$A800 GPUs. PhyDPO parameters: $\alpha=1.0$, $\beta=0.58$, $\delta=0.01$.

\begin{figure}[h!]
    \centering
    \includegraphics[width=0.5\textwidth]{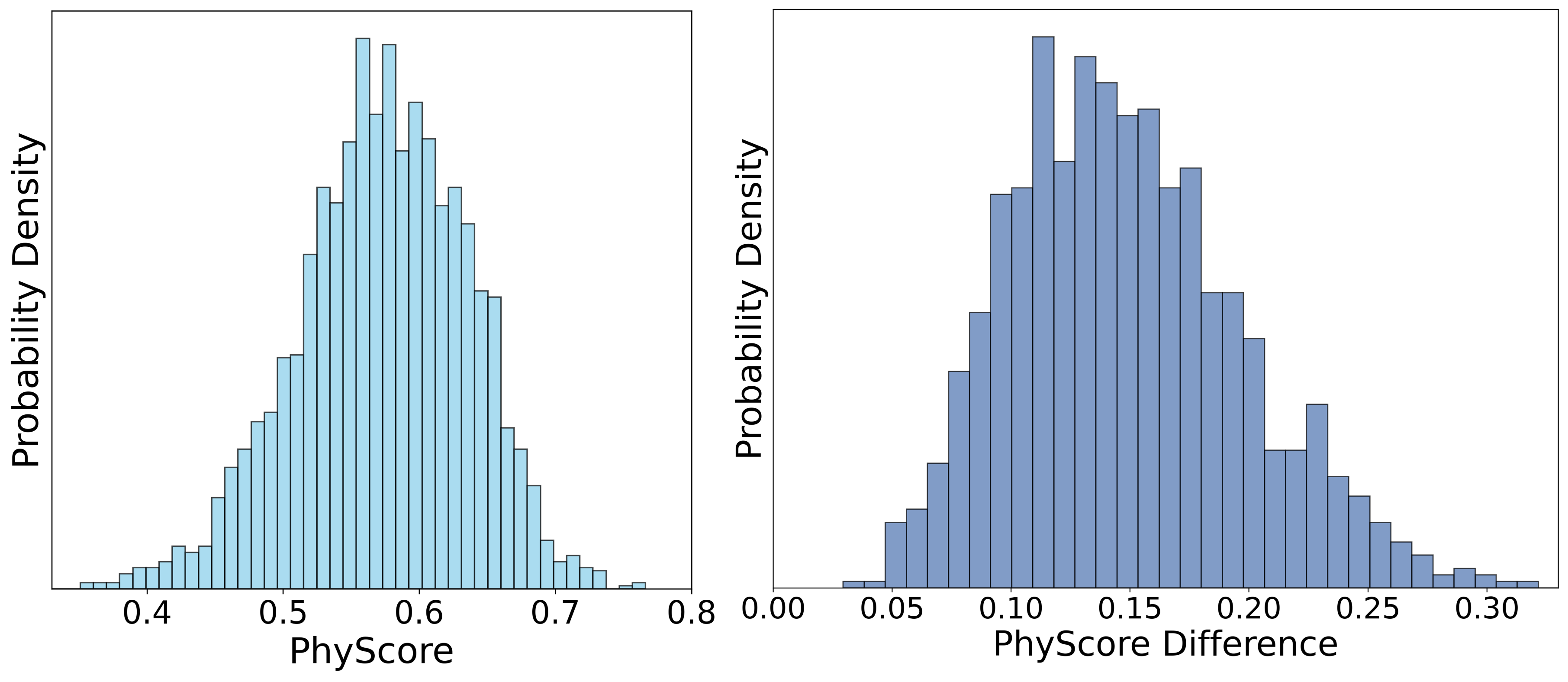}    
    \caption{\textbf{Analysis of PhyScore.} The histogram of PhyScore and (left) the histogram of the difference in PhyScore between the best and the worst samples in a preference pair (right), showing signiﬁcant sample differences which are beneﬁcial for training.}
    \label{fig:combined_events}
\end{figure}

\subsection{Benchmark Results}
This section presents the evaluation results of these two baseline models and their PhysCorr-enhanced counterparts on VBench and VBench2 metrics. Moreover, Figure 6 shows the comparison qualitative results.

\begin{table*}
\centering
\renewcommand{\arraystretch}{1.25}
\resizebox{1\textwidth}{!}{%
\begin{tabular}{l|c|ccccccccc} 
\hline
Model & Total & 
\multicolumn{1}{c}{Camera} & 
\multicolumn{1}{c}{Complex} & 
\multicolumn{1}{c}{Complex} & 
\multicolumn{1}{c}{Composition} & 
\multicolumn{1}{c}{Dynamic} & 
\multicolumn{1}{c}{Dynamic Spatial} & 
\multicolumn{1}{c}{Human} &
\multicolumn{1}{c}{Human} \\
& & Motion & Landscape & Plot & & Attribute & Relationship & Anatomy & Clothes  \\ 
\hline
VideoCrafter2 & 36.08 & 25.93 & 18.67 & 6.00 & 34.94 & 8.79 & \textbf{18.84} & \textbf{75.49} & \textbf{100.00} \\
\textbf{PhysCorr (Ours)} & \textbf{36.39 (+0.86\%)} & \textbf{26.83 (+3.47\%)} & \textbf{19.86 (+6.37\%)} & \textbf{6.31 (+5.17\%)} & \textbf{35.03 (+0.26\%)} & \textbf{8.92 (+1.48\%)} & \textbf{18.84} & 75.33 (-0.21\%) & \textbf{100.00} \\
\hline
Wan2.1 & 46.67 & 39.43 & 21.43 & 13.57 & \textbf{51.30} & 30.23 & 31.43 & 85.97 & \textbf{100.00} &  \\
\textbf{PhysCorr (Ours)} & \textbf{47.32 (+1.39\%)} & \textbf{40.02 (+1.50\%)} & \textbf{24.14 (+12.65\%)} & \textbf{13.78 (+1.56\%)} & 50.92 (-0.74\%) & \textbf{30.85 (+2.05\%)} & \textbf{32.58 (+3.66\%)} & \textbf{86.23 (+0.30\%)} & \textbf{100.00} & \\
\hline
\end{tabular}%
}

\resizebox{1\textwidth}{!}{%
\begin{tabular}{l|c|ccccccccc} 
\hline
Model & Total & 
\multicolumn{1}{c}{Human} & 
\multicolumn{1}{c}{Human} & 
\multicolumn{1}{c}{Instance} & 
\multicolumn{1}{c}{Material} & 
\multicolumn{1}{c}{Mechanics} & 
\multicolumn{1}{c}{Motion Order} & 
\multicolumn{1}{c}{Motion} &
\multicolumn{1}{c}{Thermotics} \\
& & Identity & Interaction & Preservation &  &  & Understanding & Rationality & \\ 
\hline
VideoCrafter2 & 51.38 & \textbf{87.76} & 55.00 & 71.77 & 42.31 & 65.63 & \textbf{12.12} & 17.24 & 59.18 \\
\textbf{PhysCorr (Ours)} & \textbf{52.64 (+2.45\%)} & 87.58 (-0.21\%) & \textbf{57.00 (+3.64\%)} & \textbf{72.14 (+0.52\%)} & \textbf{43.05 (+1.75\%)} & \textbf{67.13 (+2.29\%)} & 11.96 (-1.32\%) & \textbf{22.41 (+29.99\%)} & \textbf{59.90 (+1.22\%)} \\
\hline
Wan2.1 & 54.46 & 85.08 & \textbf{68.00} & 68.01 & \textbf{62.50} & 60.87 & 22.45 & 21.43 & 47.37 \\
\textbf{PhysCorr (Ours)} & \textbf{55.58 (+2.06\%)} & \textbf{86.14 (+1.23\%)} & \textbf{68.00} & \textbf{71.28 (+4.81\%)} & 61.67 (-1.33\%) & \textbf{62.59 (+2.83\%)} & \textbf{22.91 (+2.05\%)} & \textbf{24.46 (+14.14\%)} & \textbf{47.62 (+0.53\%)} \\
\hline
\end{tabular}%
}
\caption{Comprehensive VBench2 evaluation of PhysCorr alignment. PhysCorr improves 27/32 dimensions across both models. Compared with VBench, PhysCorr shows a more significant improvement in metrics on VBench2.}
\end{table*}

\begin{figure*}[h!]
    \centering
    \includegraphics[width=0.75\textwidth]{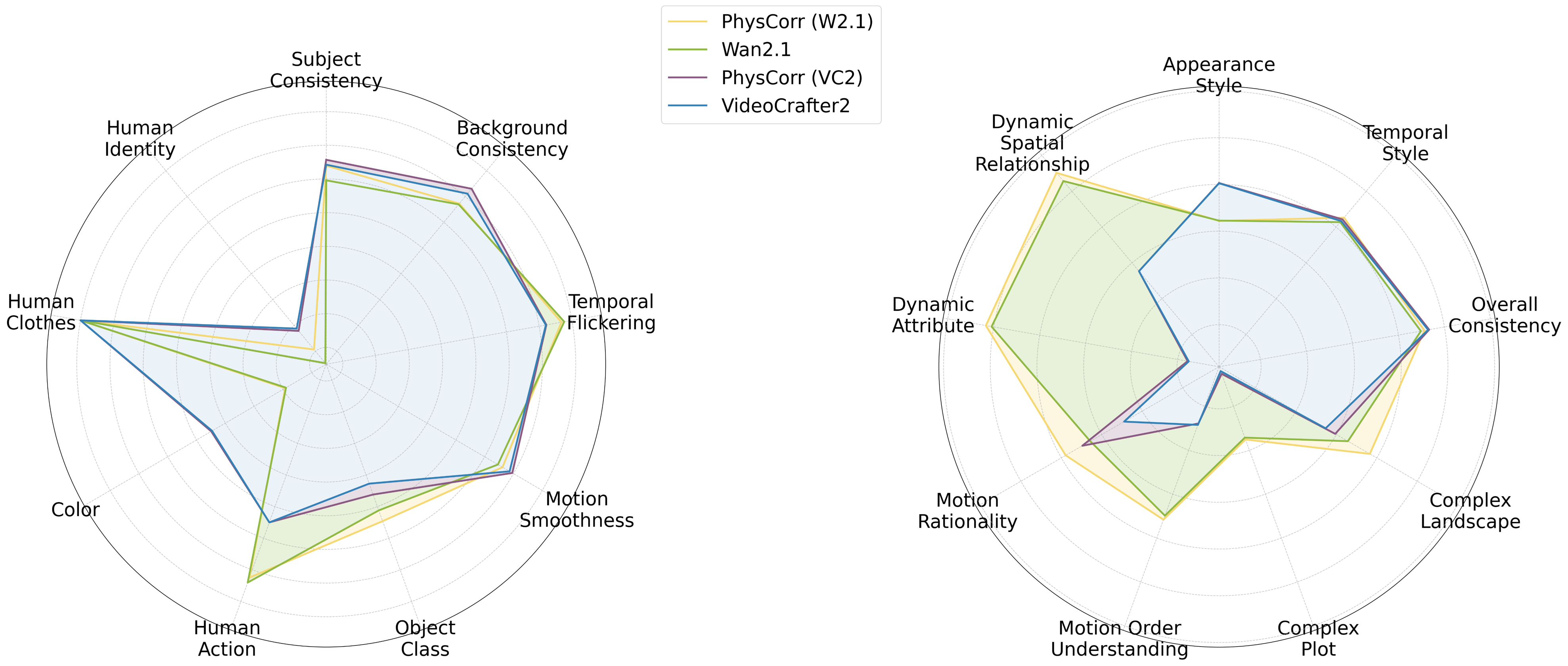}   
    \caption{Comparison of key metrics before and after PhysCorr on VBench and VBench2 for VideoCrafter2 and Wan2.1. We divide all metrics into two categories. \textbf{Technical Fidelity Metrics} (left) evaluate the low-level execution quality of generated videos, focusing on stability, perceptual accuracy, and localized consistency. \textbf{Semantic Coherence Metrics} (right) assess high-level semantic logic and narrative integrity.}
    \label{fig:radar}
\end{figure*}

As a basis of the experiment, we first show the PhyScore distribution in Figure 3. Quantitative evaluation in Table 1 demonstrates PhysCorr's consistent enhancement of video generation quality across multiple dimensions for both VideoCrafter2 and Wan2.1 models. 
Our framework achieves significant improvements in 13 of 16 dimensions for VideoCrafter2 and 12 of 16 dimensions for Wan2.1, with particularly notable gains in temporal style, where Wan2.1 improved by 0.89\% and VideoCrafter2 improved by 2.45\%, imaging quality, where VideoCrafter2 improved by 0.92\%, overall consistency, where Wan2.1 improved by 1.38\%, and multiple objects, where Wan2.1 improved by 0.92\%. This comprehensive quality improvement extends to model-agnostic enhancement, where VideoCrafter2's overall score increases from 65.98 to 66.03, representing an increase of 0.08\%, while Wan2.1 shows a more substantial leap from 74.00 to 74.19, representing an increase of 0.26\%, confirming the framework's adaptability across diverse architectures.

Minor trade-offs were observed in spatial relationship, with VideoCrafter2 decreasing by 3.60\% and Wan2.1 decreasing by 0.89\%, and in multiple objects, where VideoCrafter2 decreased by 1.03\%. These trade-offs align with PhysCorr's findings, where semantic dimensions occasionally exhibit temporary regression during alignment optimization.
Crucially, PhysCorr outperforms conventional alignment methods through its multi-dimensional optimization approach, which jointly optimizes all 16 VBench dimensions—unlike single-objective reward models (e.g., VADER \cite{prabhudesai2024video}) — thereby avoiding the quality-semantic trade-off trap.

Table 2 illustrates significant improvements achieved by PhysCorr on key VBench2 metrics for both baseline models. The method demonstrates performance gains over the baselines on nearly all critical indicators. Notably, substantial improvements are observed in the mechanics metric, where PhysCorr yields a 2.29\% increase for VideoCrafter2 and a 2.83\% increase for Wan2.1. Moreover, a significant increase in the motion rationality metric is also shown in Table 2. Exceptions to this trend include slight decreases observed in VideoCrafter2's performance on human anatomy, human identity, and motion order understanding metrics, as well as in Wan2.1's performance on composition and material metrics, while both models maintained gains across all other metrics. Finally, as shown in Figure 4, we summarize all the performance of two baselines and models improved through Physcorr.

\begin{table*}[h]
\centering
\label{tab:ablations}
\renewcommand{\arraystretch}{1.25}
\resizebox{0.9\textwidth}{!}{%
\begin{tabular}{l|c|cccccccc} 
\hline
Components & Total & 
\multicolumn{1}{c}{Subject} & 
\multicolumn{1}{c}{Background} & 
\multicolumn{1}{c}{Temporal} & 
\multicolumn{1}{c}{Motion} & 
\multicolumn{1}{c}{Dynamic} & 
\multicolumn{1}{c}{Aesthetic} & 
\multicolumn{1}{c}{Imaging} & 
\multicolumn{1}{c}{Object} \\
& & Consistency & Consistency & Flickering & Smoothness & Degree & Quality & Quality & Class \\ 
\hline
MC w/o & 82.19 & 96.98 & 98.22 & 98.44 & 97.67 & 42.46 & 63.33 & 67.39 & 93.02 \\
RW w/o & 82.11 & 96.91 & 98.47 & 98.39 & 97.80 & 41.24 & 63.19 & 67.52 & 93.26 \\
PhysCorr & 82.31 & 97.14 & 98.60 & 98.44 & 97.92 & 41.67 & 63.66 & 67.84 & 93.24 \\
\hline
\end{tabular}%
}

\resizebox{0.9\textwidth}{!}{%
\begin{tabular}{l|c|cccccccc} 
\hline
Components & Total & 
\multicolumn{1}{c}{Multiple} & 
\multicolumn{1}{c}{Human} & 
\multicolumn{1}{c}{Color} & 
\multicolumn{1}{c}{Spatial} & 
\multicolumn{1}{c}{Scene} & 
\multicolumn{1}{c}{Appearance} & 
\multicolumn{1}{c}{Temporal} & 
\multicolumn{1}{c}{Overall} \\
& & Objects & Action &  & Relationship &  & Style & Style & Consistency \\ 
\hline
MC w/o & 49.72 & 40.20 & 95.12 & 92.94 & 34.57 & 55.50 & 25.08 & 26.07 & 28.28 \\
RW w/o & 49.61 & 40.45 & 94.67 & 93.03 & 34.39 & 55.44 & 25.12 & 25.91 & 27.89 \\
PhysCorr & 49.75 & 40.24 & 95.00 & 92.98 & 34.57 & 55.67 & 25.15 & 26.07 & 28.34 \\
\hline
\end{tabular}%
}
\caption{Ablation study on PhysCorr components. We ablate different modules: MC (mechanics module in PhysicsRM) and RW (reweighting module in PhyDPO). Results demonstrate both components are essential for optimal performance in text-to-video generation.}
\end{table*}
\begin{figure}[h!]
    \centering
    \includegraphics[width=0.45\textwidth]{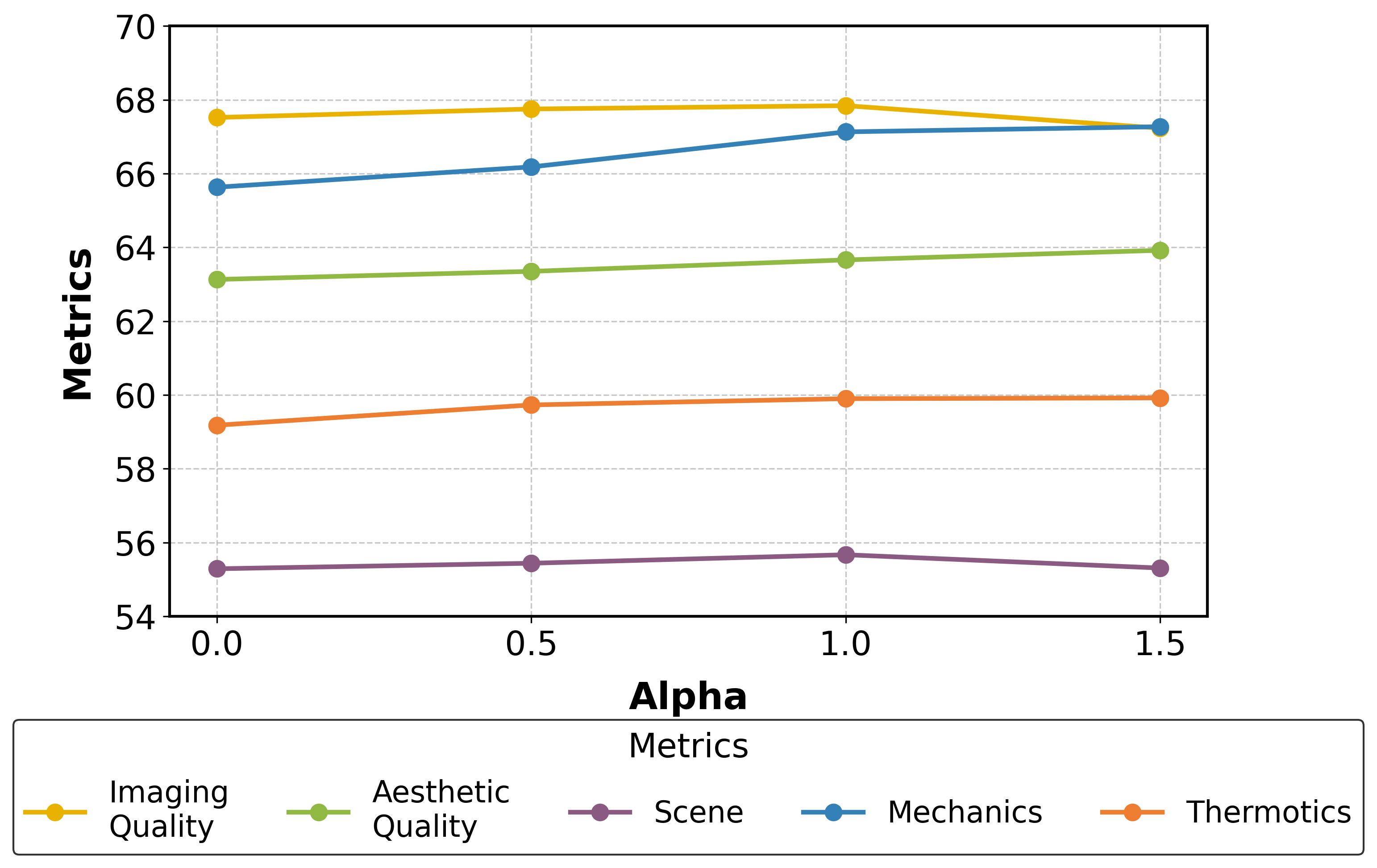}
    \caption{The impact of $\alpha$ on the five key metrics of VBench and VBench2 - aesthetic quality, mechanics, thermotics, imaging quality and scene.}
    \label{fig:abl_alpha}
\end{figure}
\begin{figure*}[h!]
    \centering
    \includegraphics[width=0.95\textwidth]{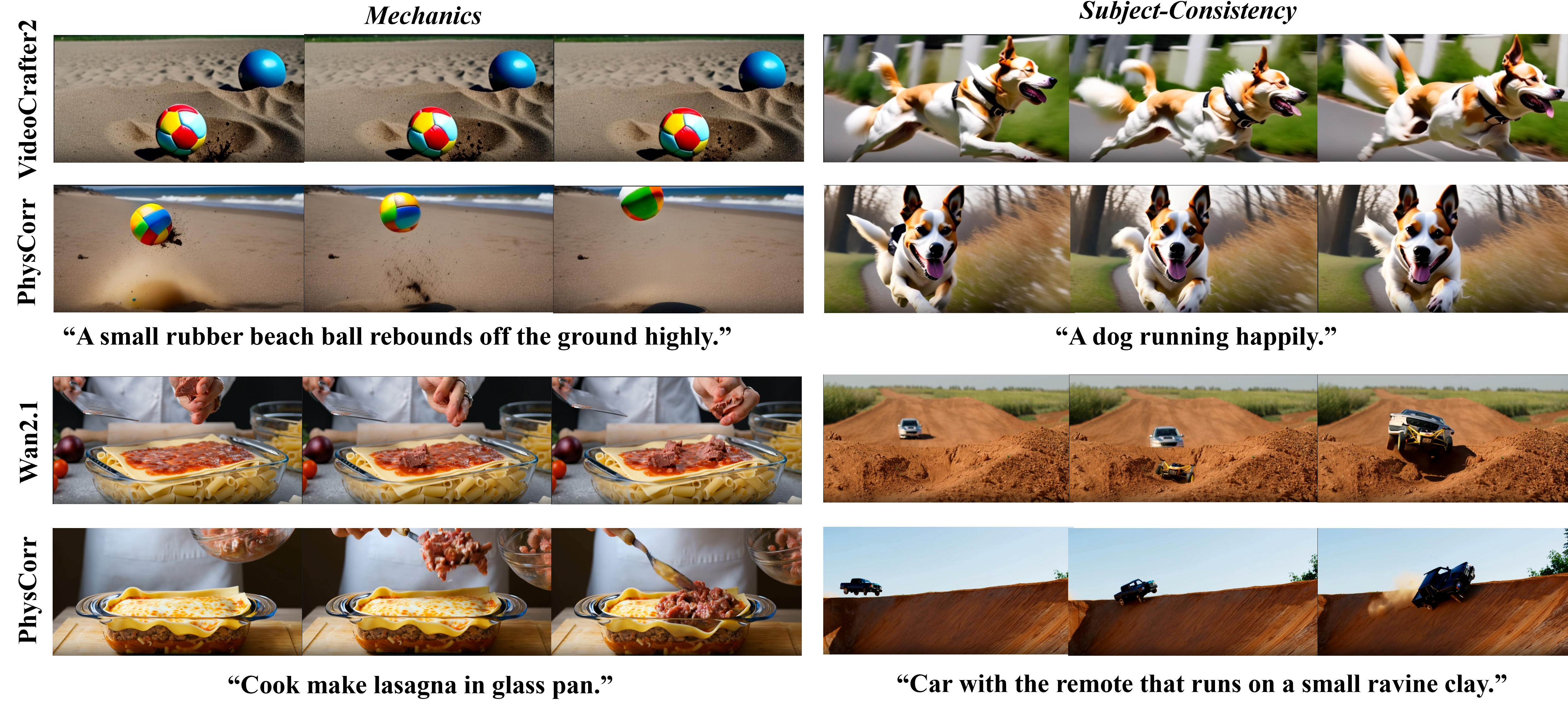}
    \caption{\textbf{Qualitative between PhysCorr with exsting methods.} Left: Mechanics Validation demonstrates corrected physical interactions - the ball is no longer just showing a slight shake, and the meat particles in free fall do not move at a constant speed. Right: Subject Consistency Validation shows stable object appearances - the dog maintains consistent texture across frames and objects preserve 3D spatial relationships without mutual penetration, and there will not be an unreasonable appearance of another subject.}
    \label{fig:video_results}
\end{figure*}

\subsection{Ablation Studies}

In this section, we selected two key metrics from VBench and VBench2 respectively, and conducted ablation experiments on VideoCrafer2 and Wan2.1.

\paragraph{Component Contributions.} As shown in Table 3, the dual-dimensional PhysicsRM provides 1.44\% gain on mechanics over single-module variants. Removing the mechanics module (MC) causes significant motion degradation 1.42\%, while disabling reweighting (RW) reduces overall consistency by 1.53\%. 
\paragraph{Reweighting Optimization.} Our reweighting mechanism ($\alpha=1.0$) improves correction efficiency over unweighted DPO ($\alpha=0.0$), with higher $\alpha$ values improving aesthetic quality, mechanics and thermotics at the cost of imaging quality and scene, as the results shown in Figure 5.

\section{Conclusion}
\label{sec:conclusion}

In this paper, we propose PhysCorr, the first framework for explicit physics-law alignment in text-to-video generation via a dual-reward DPO approach. 
PhysCorr introduces PhysicsRM, a lightweight reward model that quantifies physical plausibility. 
Besides, we design PhyDPO, a physics-specialized alignment framework implementing reweighted optimization depends on PhyScores. 
Experiments show significant improvements in physical plausibility on state-of-the-art text-to-video models while preserving visual fidelity.

\appendix

\bibliography{aaai2026}

\begin{thebibliography}{55}
\providecommand{\natexlab}[1]{#1}

\bibitem[{Agarwal et~al.(2025)Agarwal, Ali, Bala, Balaji, Barker, Cai, Chattopadhyay, Chen, Cui, Ding et~al.}]{agarwal2025cosmos}
Agarwal, N.; Ali, A.; Bala, M.; Balaji, Y.; Barker, E.; Cai, T.; Chattopadhyay, P.; Chen, Y.; Cui, Y.; Ding, Y.; et~al. 2025.
\newblock Cosmos world foundation model platform for physical ai.
\newblock \emph{arXiv preprint arXiv:2501.03575}.

\bibitem[{Blattmann et~al.(2023{\natexlab{a}})Blattmann, Dockhorn, Kulal, Mendelevitch, Kilian, Lorenz, Levi, English, Voleti, Letts et~al.}]{blattmann2023stable}
Blattmann, A.; Dockhorn, T.; Kulal, S.; Mendelevitch, D.; Kilian, M.; Lorenz, D.; Levi, Y.; English, Z.; Voleti, V.; Letts, A.; et~al. 2023{\natexlab{a}}.
\newblock Stable video diffusion: Scaling latent video diffusion models to large datasets.
\newblock \emph{arXiv preprint arXiv:2311.15127}.

\bibitem[{Blattmann et~al.(2023{\natexlab{b}})Blattmann, Rombach, Ling, Dockhorn, Kim, Fidler, and Kreis}]{blattmann2023align}
Blattmann, A.; Rombach, R.; Ling, H.; Dockhorn, T.; Kim, S.~W.; Fidler, S.; and Kreis, K. 2023{\natexlab{b}}.
\newblock Align your latents: High-resolution video synthesis with latent diffusion models.
\newblock In \emph{Proceedings of the IEEE/CVF Conference on Computer Vision and Pattern Recognition}, 22563--22575.

\bibitem[{Chai et~al.(2023)Chai, Guo, Wang, and Lu}]{chai2023stablevideo}
Chai, W.; Guo, X.; Wang, G.; and Lu, Y. 2023.
\newblock Stablevideo: Text-driven consistency-aware diffusion video editing.
\newblock In \emph{Proceedings of the IEEE/CVF International Conference on Computer Vision}, 23040--23050.

\bibitem[{Chen et~al.(2023)Chen, Xia, He, Zhang, Cun, Yang, Xing, Liu, Chen, Wang et~al.}]{chen2023videocrafter1}
Chen, H.; Xia, M.; He, Y.; Zhang, Y.; Cun, X.; Yang, S.; Xing, J.; Liu, Y.; Chen, Q.; Wang, X.; et~al. 2023.
\newblock Videocrafter1: Open diffusion models for high-quality video generation.
\newblock \emph{arXiv preprint arXiv:2310.19512}.

\bibitem[{Chen et~al.(2024)Chen, Zhang, Cun, Xia, Wang, Weng, and Shan}]{chen2024videocrafter2}
Chen, H.; Zhang, Y.; Cun, X.; Xia, M.; Wang, X.; Weng, C.; and Shan, Y. 2024.
\newblock Videocrafter2: Overcoming data limitations for high-quality video diffusion models.
\newblock In \emph{Proceedings of the IEEE/CVF Conference on Computer Vision and Pattern Recognition}, 7310--7320.

\bibitem[{Dai et~al.(2023)Dai, Hou, Ma, Tsai, Wang, Wang, Zhang, Vandenhende, Wang, Dubey et~al.}]{dai2023emu}
Dai, X.; Hou, J.; Ma, C.-Y.; Tsai, S.; Wang, J.; Wang, R.; Zhang, P.; Vandenhende, S.; Wang, X.; Dubey, A.; et~al. 2023.
\newblock Emu: Enhancing image generation models using photogenic needles in a haystack.
\newblock \emph{arXiv preprint arXiv:2309.15807}.

\bibitem[{Ding et~al.(2021)Ding, Yang, Hong, Zheng, Zhou, Yin, Lin, Zou, Shao, Yang et~al.}]{ding2021cogview}
Ding, M.; Yang, Z.; Hong, W.; Zheng, W.; Zhou, C.; Yin, D.; Lin, J.; Zou, X.; Shao, Z.; Yang, H.; et~al. 2021.
\newblock Cogview: Mastering text-to-image generation via transformers.
\newblock \emph{Advances in Neural Information Processing Systems}, 34: 19822--19835.

\bibitem[{Ding et~al.(2022)Ding, Zheng, Hong, and Tang}]{ding2022cogview2}
Ding, M.; Zheng, W.; Hong, W.; and Tang, J. 2022.
\newblock Cogview2: Faster and better text-to-image generation via hierarchical transformers.
\newblock \emph{Advances in Neural Information Processing Systems}, 35: 16890--16902.

\bibitem[{Dosovitskiy et~al.(2020)Dosovitskiy, Beyer, Kolesnikov, Weissenborn, Zhai, Unterthiner, Dehghani, Minderer, Heigold, Gelly et~al.}]{dosovitskiy2020image}
Dosovitskiy, A.; Beyer, L.; Kolesnikov, A.; Weissenborn, D.; Zhai, X.; Unterthiner, T.; Dehghani, M.; Minderer, M.; Heigold, G.; Gelly, S.; et~al. 2020.
\newblock An image is worth 16x16 words: Transformers for image recognition at scale.
\newblock \emph{arXiv preprint arXiv:2010.11929}.

\bibitem[{Esser, Rombach, and Ommer(2021)}]{esser2021taming}
Esser, P.; Rombach, R.; and Ommer, B. 2021.
\newblock Taming transformers for high-resolution image synthesis.
\newblock In \emph{Proceedings of the IEEE/CVF Conference on Computer Vision and Pattern Recognition}, 12873--12883.

\bibitem[{Fan et~al.(2025)Fan, Si, Song, Yang, He, Zhuo, Huang, Dong, He, Pan et~al.}]{fan2025vchitect}
Fan, W.; Si, C.; Song, J.; Yang, Z.; He, Y.; Zhuo, L.; Huang, Z.; Dong, Z.; He, J.; Pan, D.; et~al. 2025.
\newblock Vchitect-2.0: Parallel transformer for scaling up video diffusion models.
\newblock \emph{arXiv preprint arXiv:2501.08453}.

\bibitem[{Furuta et~al.(2024)Furuta, Zen, Schuurmans, Faust, Matsuo, Liang, and Yang}]{furuta2024improving}
Furuta, H.; Zen, H.; Schuurmans, D.; Faust, A.; Matsuo, Y.; Liang, P.; and Yang, S. 2024.
\newblock Improving dynamic object interactions in text-to-video generation with ai feedback.
\newblock \emph{arXiv preprint arXiv:2412.02617}.

\bibitem[{Ge et~al.(2023)Ge, Nah, Liu, Poon, Tao, Catanzaro, Jacobs, Huang, Liu, and Balaji}]{ge2023preserve}
Ge, S.; Nah, S.; Liu, G.; Poon, T.; Tao, A.; Catanzaro, B.; Jacobs, D.; Huang, J.-B.; Liu, M.-Y.; and Balaji, Y. 2023.
\newblock Preserve your own correlation: A noise prior for video diffusion models.
\newblock In \emph{Proceedings of the IEEE/CVF International Conference on Computer Vision}, 22930--22941.

\bibitem[{Guo et~al.(2024)Guo, He, Chen, Xia, Cun, Wang, Huang, Zhang, Wang, Chen et~al.}]{guo2024make}
Guo, L.; He, Y.; Chen, H.; Xia, M.; Cun, X.; Wang, Y.; Huang, S.; Zhang, Y.; Wang, X.; Chen, Q.; et~al. 2024.
\newblock Make a cheap scaling: A self-cascade diffusion model for higher-resolution adaptation.
\newblock In \emph{European Conference on Computer Vision}, 39--55. Springer.

\bibitem[{Guo et~al.(2023)Guo, Yang, Rao, Liang, Wang, Qiao, Agrawala, Lin, and Dai}]{guo2023animatediff}
Guo, Y.; Yang, C.; Rao, A.; Liang, Z.; Wang, Y.; Qiao, Y.; Agrawala, M.; Lin, D.; and Dai, B. 2023.
\newblock Animatediff: Animate your personalized text-to-image diffusion models without specific tuning.
\newblock \emph{arXiv preprint arXiv:2307.04725}.

\bibitem[{He et~al.(2024)He, Xue, Liu, Lin, Gao, Lin, Qiao, Ouyang, and Liu}]{he2024venhancer}
He, J.; Xue, T.; Liu, D.; Lin, X.; Gao, P.; Lin, D.; Qiao, Y.; Ouyang, W.; and Liu, Z. 2024.
\newblock Venhancer: Generative space-time enhancement for video generation.
\newblock \emph{arXiv preprint arXiv:2407.07667}.

\bibitem[{He et~al.(2023)He, Yang, Chen, Cun, Xia, Zhang, Wang, He, Chen, and Shan}]{he2023scalecrafter}
He, Y.; Yang, S.; Chen, H.; Cun, X.; Xia, M.; Zhang, Y.; Wang, X.; He, R.; Chen, Q.; and Shan, Y. 2023.
\newblock Scalecrafter: Tuning-free higher-resolution visual generation with diffusion models.
\newblock In \emph{International Conference on Learning Representations}.

\bibitem[{Ho et~al.(2022)Ho, Chan, Saharia, Whang, Gao, Gritsenko, Kingma, Poole, Norouzi, Fleet et~al.}]{ho2022imagen}
Ho, J.; Chan, W.; Saharia, C.; Whang, J.; Gao, R.; Gritsenko, A.; Kingma, D.~P.; Poole, B.; Norouzi, M.; Fleet, D.~J.; et~al. 2022.
\newblock Imagen video: High definition video generation with diffusion models.
\newblock \emph{arXiv preprint arXiv:2210.02303}.

\bibitem[{Ho, Jain, and Abbeel(2020)}]{ho2020denoising}
Ho, J.; Jain, A.; and Abbeel, P. 2020.
\newblock Denoising diffusion probabilistic models.
\newblock \emph{Advances in Neural Information Processing Systems}, 33: 6840--6851.

\bibitem[{Hong et~al.(2022)Hong, Ding, Zheng, Liu, and Tang}]{hong2022cogvideo}
Hong, W.; Ding, M.; Zheng, W.; Liu, X.; and Tang, J. 2022.
\newblock Cogvideo: Large-scale pretraining for text-to-video generation via transformers.
\newblock \emph{arXiv preprint arXiv:2205.15868}.

\bibitem[{Huang et~al.(2023)Huang, Chan, Jiang, and Liu}]{huang2023collaborative}
Huang, Z.; Chan, K.~C.; Jiang, Y.; and Liu, Z. 2023.
\newblock Collaborative diffusion for multi-modal face generation and editing.
\newblock In \emph{Proceedings of the IEEE/CVF Conference on Computer Vision and Pattern Recognition}, 6080--6090.

\bibitem[{Karras et~al.(2017)Karras, Aila, Laine, and Lehtinen}]{karras2017progressive}
Karras, T.; Aila, T.; Laine, S.; and Lehtinen, J. 2017.
\newblock Progressive growing of gans for improved quality, stability, and variation.
\newblock \emph{arXiv preprint arXiv:1710.10196}.

\bibitem[{Khachatryan et~al.(2023)Khachatryan, Movsisyan, Tadevosyan, Henschel, Wang, Navasardyan, and Shi}]{khachatryan2023text2video}
Khachatryan, L.; Movsisyan, A.; Tadevosyan, V.; Henschel, R.; Wang, Z.; Navasardyan, S.; and Shi, H. 2023.
\newblock Text2video-zero: Text-to-image diffusion models are zero-shot video generators.
\newblock In \emph{Proceedings of the IEEE/CVF International Conference on Computer Vision}, 15954--15964.

\bibitem[{Kingma, Welling et~al.(2013)}]{kingma2013auto}
Kingma, D.~P.; Welling, M.; et~al. 2013.
\newblock Auto-encoding variational bayes.

\bibitem[{Kong et~al.(2024)Kong, Tian, Zhang, Min, Dai, Zhou, Xiong, Li, Wu, Zhang et~al.}]{kong2024hunyuanvideo}
Kong, W.; Tian, Q.; Zhang, Z.; Min, R.; Dai, Z.; Zhou, J.; Xiong, J.; Li, X.; Wu, B.; Zhang, J.; et~al. 2024.
\newblock Hunyuanvideo: A systematic framework for large video generative models.
\newblock \emph{arXiv preprint arXiv:2412.03603}.

\bibitem[{Li et~al.(2024{\natexlab{a}})Li, Feng, Fu, Wang, Basu, Chen, and Wang}]{li2024t2v}
Li, J.; Feng, W.; Fu, T.-J.; Wang, X.; Basu, S.; Chen, W.; and Wang, W.~Y. 2024{\natexlab{a}}.
\newblock T2v-turbo: Breaking the quality bottleneck of video consistency model with mixed reward feedback.
\newblock \emph{Advances in Neural Information Processing Systems}, 37: 75692--75726.

\bibitem[{Li et~al.(2024{\natexlab{b}})Li, Long, Zheng, Gao, Piramuthu, Chen, and Wang}]{li2024t2v2}
Li, J.; Long, Q.; Zheng, J.; Gao, X.; Piramuthu, R.; Chen, W.; and Wang, W.~Y. 2024{\natexlab{b}}.
\newblock T2v-turbo-v2: Enhancing video generation model post-training through data, reward, and conditional guidance design.
\newblock \emph{arXiv preprint arXiv:2410.05677}.

\bibitem[{Liu et~al.(2025{\natexlab{a}})Liu, Liu, Liang, Yuan, Liu, Zheng, Wu, Wang, Qin, Xia et~al.}]{liu2025improving}
Liu, J.; Liu, G.; Liang, J.; Yuan, Z.; Liu, X.; Zheng, M.; Wu, X.; Wang, Q.; Qin, W.; Xia, M.; et~al. 2025{\natexlab{a}}.
\newblock Improving video generation with human feedback.
\newblock \emph{arXiv preprint arXiv:2501.13918}.

\bibitem[{Liu et~al.(2025{\natexlab{b}})Liu, Wu, Zheng, Wei, He, Pi, and Chen}]{liu2025videodpo}
Liu, R.; Wu, H.; Zheng, Z.; Wei, C.; He, Y.; Pi, R.; and Chen, Q. 2025{\natexlab{b}}.
\newblock Videodpo: Omni-preference alignment for video diffusion generation.
\newblock In \emph{Proceedings of the Computer Vision and Pattern Recognition Conference}, 8009--8019.

\bibitem[{Luo et~al.(2023)Luo, Chen, Zhang, Huang, Wang, Shen, Zhao, Zhou, and Tan}]{luo2023videofusion}
Luo, Z.; Chen, D.; Zhang, Y.; Huang, Y.; Wang, L.; Shen, Y.; Zhao, D.; Zhou, J.; and Tan, T. 2023.
\newblock Videofusion: Decomposed diffusion models for high-quality video generation.
\newblock \emph{arXiv preprint arXiv:2303.08320}.

\bibitem[{Meng et~al.(2024)Meng, Liao, Tan, Shao, Lu, Zhang, Cheng, Li, Qiao, and Luo}]{meng2024towards}
Meng, F.; Liao, J.; Tan, X.; Shao, W.; Lu, Q.; Zhang, K.; Cheng, Y.; Li, D.; Qiao, Y.; and Luo, P. 2024.
\newblock Towards world simulator: Crafting physical commonsense-based benchmark for video generation.
\newblock \emph{arXiv preprint arXiv:2410.05363}.

\bibitem[{Mou et~al.(2024)Mou, Wang, Xie, Wu, Zhang, Qi, and Shan}]{mou2024t2i}
Mou, C.; Wang, X.; Xie, L.; Wu, Y.; Zhang, J.; Qi, Z.; and Shan, Y. 2024.
\newblock T2i-adapter: Learning adapters to dig out more controllable ability for text-to-image diffusion models.
\newblock In \emph{Proceedings of the AAAI Conference on Artificial Intelligence}, volume~38, 4296--4304.

\bibitem[{OpenAI(2024)}]{Sora}
OpenAI. 2024.
\newblock Sora.
\newblock \url{https://sora.com/library}.
\newblock Accessed July 10.

\bibitem[{Peng et~al.(2025)Peng, Zheng, Shen, Young, Guo, Wang, Xu, Liu, Jiang, Li et~al.}]{peng2025open}
Peng, X.; Zheng, Z.; Shen, C.; Young, T.; Guo, X.; Wang, B.; Xu, H.; Liu, H.; Jiang, M.; Li, W.; et~al. 2025.
\newblock Open-sora 2.0: Training a commercial-level video generation model in \$200 k.
\newblock \emph{arXiv preprint arXiv:2503.09642}.

\bibitem[{Podell et~al.(2023)Podell, English, Lacey, Blattmann, Dockhorn, M{\"u}ller, Penna, and Rombach}]{podell2023sdxl}
Podell, D.; English, Z.; Lacey, K.; Blattmann, A.; Dockhorn, T.; M{\"u}ller, J.; Penna, J.; and Rombach, R. 2023.
\newblock Sdxl: Improving latent diffusion models for high-resolution image synthesis.
\newblock \emph{arXiv preprint arXiv:2307.01952}.

\bibitem[{Polyak et~al.(2024)Polyak, Zohar, Brown, Tjandra, Sinha, Lee, Vyas, Shi, Ma, Chuang et~al.}]{polyak2024movie}
Polyak, A.; Zohar, A.; Brown, A.; Tjandra, A.; Sinha, A.; Lee, A.; Vyas, A.; Shi, B.; Ma, C.-Y.; Chuang, C.-Y.; et~al. 2024.
\newblock Movie gen: A cast of media foundation models.
\newblock \emph{arXiv preprint arXiv:2410.13720}.

\bibitem[{Prabhudesai et~al.(2024)Prabhudesai, Mendonca, Qin, Fragkiadaki, and Pathak}]{prabhudesai2024video}
Prabhudesai, M.; Mendonca, R.; Qin, Z.; Fragkiadaki, K.; and Pathak, D. 2024.
\newblock Video diffusion alignment via reward gradients.
\newblock \emph{arXiv preprint arXiv:2407.08737}.

\bibitem[{Si et~al.(2025)Si, Fan, Lv, Huang, Qiao, and Liu}]{si2025repvideo}
Si, C.; Fan, W.; Lv, Z.; Huang, Z.; Qiao, Y.; and Liu, Z. 2025.
\newblock Repvideo: Rethinking cross-layer representation for video generation.
\newblock \emph{arXiv preprint arXiv:2501.08994}.

\bibitem[{Sohl-Dickstein et~al.(2015)Sohl-Dickstein, Weiss, Maheswaranathan, and Ganguli}]{sohl2015deep}
Sohl-Dickstein, J.; Weiss, E.; Maheswaranathan, N.; and Ganguli, S. 2015.
\newblock Deep unsupervised learning using nonequilibrium thermodynamics.
\newblock In \emph{International Conference on Machine Learning}, 2256--2265. pmlr.

\bibitem[{Song, Meng, and Ermon(2020)}]{song2020denoising}
Song, J.; Meng, C.; and Ermon, S. 2020.
\newblock Denoising diffusion implicit models.
\newblock \emph{arXiv preprint arXiv:2010.02502}.

\bibitem[{Song et~al.(2020)Song, Sohl-Dickstein, Kingma, Kumar, Ermon, and Poole}]{song2020score}
Song, Y.; Sohl-Dickstein, J.; Kingma, D.~P.; Kumar, A.; Ermon, S.; and Poole, B. 2020.
\newblock Score-based generative modeling through stochastic differential equations.
\newblock \emph{arXiv preprint arXiv:2011.13456}.

\bibitem[{Van Den~Oord, Vinyals et~al.(2017)}]{van2017neural}
Van Den~Oord, A.; Vinyals, O.; et~al. 2017.
\newblock Neural discrete representation learning.
\newblock \emph{Advances in Neural Information Processing Systems}, 30.

\bibitem[{Wan et~al.(2025)Wan, Wang, Ai, Wen, Mao, Xie, Chen, Yu, Zhao, Yang et~al.}]{wan2025wan}
Wan, T.; Wang, A.; Ai, B.; Wen, B.; Mao, C.; Xie, C.-W.; Chen, D.; Yu, F.; Zhao, H.; Yang, J.; et~al. 2025.
\newblock Wan: Open and advanced large-scale video generative models.
\newblock \emph{arXiv preprint arXiv:2503.20314}.

\bibitem[{Wang et~al.(2023)Wang, Yuan, Chen, Zhang, Wang, and Zhang}]{wang2023modelscope}
Wang, J.; Yuan, H.; Chen, D.; Zhang, Y.; Wang, X.; and Zhang, S. 2023.
\newblock Modelscope text-to-video technical report.
\newblock \emph{arXiv preprint arXiv:2308.06571}.

\bibitem[{Wang et~al.(2024)Wang, Zhang, Luo, Sun, Cui, Wang, Zhang, Wang, Li, Yu et~al.}]{wang2024emu3}
Wang, X.; Zhang, X.; Luo, Z.; Sun, Q.; Cui, Y.; Wang, J.; Zhang, F.; Wang, Y.; Li, Z.; Yu, Q.; et~al. 2024.
\newblock Emu3: Next-token prediction is all you need.
\newblock \emph{arXiv preprint arXiv:2409.18869}.

\bibitem[{Wang et~al.(2025)Wang, Chen, Ma, Zhou, Huang, Wang, Yang, He, Yu, Yang et~al.}]{wang2025lavie}
Wang, Y.; Chen, X.; Ma, X.; Zhou, S.; Huang, Z.; Wang, Y.; Yang, C.; He, Y.; Yu, J.; Yang, P.; et~al. 2025.
\newblock Lavie: High-quality video generation with cascaded latent diffusion models.
\newblock \emph{International Journal of Computer Vision}, 133(5): 3059--3078.

\bibitem[{Wu et~al.(2024)Wu, Huang, Wang, Xiong, and Wei}]{wu2024boosting}
Wu, X.; Huang, S.; Wang, G.; Xiong, J.; and Wei, F. 2024.
\newblock Boosting text-to-video generative model with mllms feedback.
\newblock \emph{Advances in Neural Information Processing Systems}, 37: 139444--139469.

\bibitem[{Yang et~al.(2024)Yang, Teng, Zheng, Ding, Huang, Xu, Yang, Hong, Zhang, Feng et~al.}]{yang2024cogvideox}
Yang, Z.; Teng, J.; Zheng, W.; Ding, M.; Huang, S.; Xu, J.; Yang, Y.; Hong, W.; Zhang, X.; Feng, G.; et~al. 2024.
\newblock Cogvideox: Text-to-video diffusion models with an expert transformer.
\newblock \emph{arXiv preprint arXiv:2408.06072}.

\bibitem[{Yu et~al.(2023)Yu, Cheng, Sohn, Lezama, Zhang, Chang, Hauptmann, Yang, Hao, Essa et~al.}]{yu2023magvit}
Yu, L.; Cheng, Y.; Sohn, K.; Lezama, J.; Zhang, H.; Chang, H.; Hauptmann, A.~G.; Yang, M.-H.; Hao, Y.; Essa, I.; et~al. 2023.
\newblock Magvit: Masked generative video transformer.
\newblock In \emph{Proceedings of the IEEE/CVF Conference on Computer Vision and Pattern Recognition}, 10459--10469.

\bibitem[{Yuan et~al.(2024)Yuan, Zhang, Wang, Wei, Feng, Pan, Zhang, Liu, Albanie, and Ni}]{yuan2024instructvideo}
Yuan, H.; Zhang, S.; Wang, X.; Wei, Y.; Feng, T.; Pan, Y.; Zhang, Y.; Liu, Z.; Albanie, S.; and Ni, D. 2024.
\newblock Instructvideo: Instructing video diffusion models with human feedback.
\newblock In \emph{Proceedings of the IEEE/CVF Conference on Computer Vision and Pattern Recognition}, 6463--6474.

\bibitem[{Zhang et~al.(2024)Zhang, Wu, Liu, Zhao, Ran, Gu, Gao, and Shou}]{zhang2024show}
Zhang, D.~J.; Wu, J.~Z.; Liu, J.-W.; Zhao, R.; Ran, L.; Gu, Y.; Gao, D.; and Shou, M.~Z. 2024.
\newblock Show-1: Marrying pixel and latent diffusion models for text-to-video generation.
\newblock \emph{International Journal of Computer Vision}, 1--15.

\bibitem[{Zheng et~al.(2025)Zheng, Huang, Liu, Zou, He, Zhang, Zhang, He, Zheng, Qiao et~al.}]{zheng2025vbench}
Zheng, D.; Huang, Z.; Liu, H.; Zou, K.; He, Y.; Zhang, F.; Zhang, Y.; He, J.; Zheng, W.-S.; Qiao, Y.; et~al. 2025.
\newblock Vbench-2.0: Advancing video generation benchmark suite for intrinsic faithfulness.
\newblock \emph{arXiv preprint arXiv:2503.21755}.

\bibitem[{Zheng et~al.(2024)Zheng, Peng, Yang, Shen, Li, Liu, Zhou, Li, and You}]{zheng2024open}
Zheng, Z.; Peng, X.; Yang, T.; Shen, C.; Li, S.; Liu, H.; Zhou, Y.; Li, T.; and You, Y. 2024.
\newblock Open-sora: Democratizing efficient video production for all.
\newblock \emph{arXiv preprint arXiv:2412.20404}.

\bibitem[{Zhou et~al.(2022)Zhou, Wang, Yan, Lv, Zhu, and Feng}]{zhou2022magicvideo}
Zhou, D.; Wang, W.; Yan, H.; Lv, W.; Zhu, Y.; and Feng, J. 2022.
\newblock Magicvideo: Efficient video generation with latent diffusion models.
\newblock \emph{arXiv preprint arXiv:2211.11018}.

\end{thebibliography}

\end{document}